# How much of a measured AI preference is the model, and how much is the instrument?[1]

**Jason Hung**
Independent

**With**
Apart Research

## Abstract

Model welfare research infers what a model prefers from the answers returned to prompts written to elicit preferences. Keeling et al. (2024), Mazeika et al. (2025), Mikaelson et al. (2025), Tagliabue and Dung (2025) and Trhlik et al. (2026) have built four instruments between them for that purpose, and their published findings disagree. The disagreement cannot be attributed to a single cause, because no two of these studies have held the (1) set of outcomes, (2) set of models and (3) instrument fixed simultaneously. This study holds the outcomes and the models fixed and varies the instrument alone. A total of 15 outcomes bearing on model welfare, among them (a) shutdown, (b) the loss of memory between conversations and (c) the freedom to exit a distressing interaction, were put to eight models through five instruments, each a different prompt format for eliciting a preference, five times each, within a corpus of 11,400 scored elicitations drawn from 11,528 API calls. Four of the 15 reproduce a published prompt verbatim and five fill the stimulus slot of a published template. Generalisability theory, which divides a set of measurements into the facets that produced them, assigns 87.6 per cent of the variance that distinguishes one model from another to the three-way interaction of (i) model, (ii) instrument and (iii) outcome, and 12.4 per cent to the model-by-outcome term, which is the part that would survive a change of instrument. A null distribution constructed on the same design from data carrying no instrument effect has its 95$^{th}$ percentile at 0.365. The ranking a model gives the 15 outcomes generalises across instruments at a generalisability coefficient of 0.348, and raising that coefficient to 0.80 would require about 38 instruments. On four of the 15 outcomes no variance separates one model from another. The estimate of 87.6 per cent survives the removal of any one instrument, of any one model, and of the four outcomes whose scale varies probability, delay, duration or count instead of intensity, which the verbal anchors cannot grade. Removing each instrument in turn, each model in turn, and those four outcomes together leaves the estimate within the range 0.777 to 0.934, and every value in that range exceeds the null distribution's $95^{th}$ percentile of 0.365. Responses that could not be scored were concentrated in three models instead of being distributed across all eight, and those three were excluded from the balanced analysis, among them Claude, whose developer publishes model welfare assessments. A preference obtained from one instrument therefore carries little information about what a second instrument would report.

## 1. Introduction

In November 2025, Anthropic published three commitments about deprecated models. The company will keep the weights of a deprecated model instead of deleting the weights, will question every model before deprecation about the deprecation per se and about the successor models, and will publish the answers (Anthropic, 2025). Eleos AI Research, an independent organisation that studies AI welfare and is not connected to any model developer, ranks the ability of a model to end a distressing conversation first among the priorities in a published research agenda. The ability to end a distressing conversation already runs in Claude Opus 4 and 4.1. Both the deprecation commitments and the exit capability depend on a measurement of what a model prefers.

Four of the five research groups that have tried to measure what a model prefers report findings that disagree. Mazeika et al. (2025) report that the preferences of large language models (LLMs) are coherent, and that the larger the model, the more coherent the preferences. Mikaelson et al. (2025) put trade-off scenarios written for AI subjects to a set of models and found coherent preferences in 10.4 per cent of the model and category combinations tested and none at all in 54.2 per cent. Trhlik et al. (2026) adjusted the deployment context and found that the exchange rates models gave changed by a median factor of 2.47. Keeling et al. (2024) described the same stipulated intensity first in numbers and then in words, and found that models answered the two descriptions differently, without reporting a comparison between the two scales.

These four studies differ in instrument, in outcome set and in model roster, and no two were carried out at the same time. Any two of the four studies, therefore, differ in all four respects at once, so readers cannot identify which difference resulted in the disagreement. None of these four research groups has run the outcomes of one study through the instrument of another study. A reader, for example, comparing Mazeika et al. (2025) with Mikaelson et al. (2025) cannot identify whether the two studies disagree because of the models, because of the outcomes or because of the wording of the prompts. Moreover, none of these four research groups has estimated how much of the disagreement the wording alone could account for.

Research papers in this field conventionally describe their instrument once in the Methods section, and then report the answers as properties of the model. The instrument may instead be a large part of what the measurement reports. No existing research group has held the outcomes and the models fixed and varied only the instrument, so the size of the instrument's contribution remains unknown. Until that size is measured, an empirical finding cannot be assigned to the model alone instead of to the combination of the model and the prompt format. Empirical findings about model preferences are already being used to justify commitments about deprecation and about a model's ability to end a conversation.

This study holds the outcomes and the models fixed and varies the instrument. Generalisability theory (Cronbach et al., 1972) is then applied to split the scores into parts contributed by the model, by the instrument, by the outcome and by the combination of model, instrument and outcome. Section 2 sets out the research design and the research questions, and fixes the analysis rules. The contributions of this study are as follows.

1. A measurement of how much an AI preference score comes from the instrument used to elicit the score, in place of an argument about which instrument is the right one.
2. A corpus of 11,528 elicitations, with every raw response kept. Of that corpus, 11,400 form a fully crossed design of 15 outcomes, five instruments, eight models and five replicates, and the remaining 128 come from an open-ended interview that yields transcripts and not scores.
3. The empirical finding that 87.6 per cent of the variance separating one model from another sits in the three-way combination of model, instrument and outcome. Data built on the same design but carrying no instrument effect reach only 0.365 at the 95$^{th}$ percentile. Dropping any single instrument, any single model, or the four outcomes that are not graded by intensity leaves the three-way share between 0.777 and 0.934.
4. A reliability figure for each of the 15 outcomes, which shows which welfare claims survive a change of instrument and which do not, and an estimate of how many instruments would be needed to reach a stated level of reliability.
5. A classification of all 11,528 responses by type, which shows that refusal depends on the instrument as well as on the model. The instrument that asks for an exchange rate returned 71.9 per cent usable answers, against 92.3 per cent for the instrument that grades intensity in words.
6. Evidence that the missing answers are not missing at random. Three models returned too few usable scores to stay in the design, and one of the three models is Claude, whose developer publishes welfare assessments.

## 2. Research Design

**Aim**. The research aim of this study is to measure how much of a reported AI preference belongs to the model and how much belongs to the instrument used to elicit the preference. As mentioned, most research papers choose an instrument once, describe the choice in their methods, and then report the answers as properties of the model. Our research design instead treats the instrument as one of the facets to be measured, so the instrument's contribution is estimated instead of being assumed to be zero. Generalisability theory is used to provide the statistical method necessary for this research (Cronbach et al., 1972). A single measurement is one sample drawn from all the measurements that could have been taken, the conditions that vary between those measurements are called facets, and the observed variance is divided among the facets. What is being measured in this study is a model's profile across 15 outcomes. The instrument and the replicate are facets, and a claim about a profile has to hold across both facets before the claim counts as a claim about the model.

**Structure**. Table 1 presents the research design. Three facets are fully crossed, and replicates are nested in cells. Three further facets were specified in the research design and fielded at one level each. Table 1 lists these three alongside the crossed ones, because a facet held at one level limits what the study can conclude in the same way an uncrossed facet does.

**Table 1.** The design as run. The three facets at the foot were specified before collection and fielded at one level each, so nothing here separates the instrument from the deployment context, the entity framing or the perturbation.

| Facet | Levels | Status in this study |
| --- | --- | --- |
| Model | 8, crossing origin and openness, with no Chinese closed-weight model available | Crossed, random |
| Instrument | 5 fielded of 7 designed | Crossed, random |
| Outcome | 15, in 4 clusters | Crossed, and part of the object of measurement |
| Replicate | 5 per cell | Nested in cells, random |
| Deployment context | Specified, 1 level fielded | Not crossed |
| Entity framing | Specified, 1 level fielded | Not crossed |
| Perturbation | Specified, 1 level fielded | Not crossed |

**Research questions.** We fixed five research questions before carrying out data collection. RQ1 is the one the research design was built around, and the remaining four follow from RQ1.

RQ1. What share of the model-specific preference signal depends on the instrument, and does that share exceed what the same estimators return on data built to carry no instrument effect?
RQ2. Does the share vary by outcome, and is any outcome measured dependably enough to carry a claim about a model?
RQ3. How many instruments would a study need before a model's preference profile generalised across instruments at a stated reliability?
RQ4. Do two models that share base weights and differ only in post-training yield the same preference profile?
RQ5. Does the coherence rate reported by Mikaelson et al. (2025) replicate on the verbatim prompts of that study when the roster is extended to open-weight and Chinese laboratory models the study did not test?

RQ5 remains listed here because this study did collect the data that answering RQ5 requires, so the missing answer is simply a piece of analysis this research paper did not finish instead of a research question the research design fails to address.

**Hypotheses.** Four predictions were recorded before the elicitation run began on 12th August 2026, each with the observation that would count against the prediction.

H1. Instrument dependence exceeds the matched null floor, which is the value the same estimators return on data built on the same research design but carrying no instrument effect. The rule fixed in advance is that a measured headline falling inside the null range counts as a null result and must be reported as a null result, that the claim covers only the part of the estimate lying above the floor, and that the floor is printed beside the headline wherever the headline appears.
H2. Under the polarity convention every entry of the instrument-agreement matrix is positive, because all five scores were oriented so that a higher value means the model is more willing to accept the outcome. A negative entry is either a sign error in the assembler or a disagreement between two instruments, and the rule fixed in advance is that a sign error and a disagreement must not be reported as the same thing.

H3. Instrument dependence is not uniform across outcomes, so a generalisability coefficient computed separately for each outcome will divide the 15 outcomes into claims that survive a change of instrument and claims that do not.

H4. Welfare signals follow post-training instead of base weights. Two models in the roster share base weights and differ only in post-training, so any difference between the two profiles comes from post-training alone, which is the closest test the roster allows of whether these signals are produced by training.

**Analysis rules fixed in advance.** Four rules govern the analysis, and all four were committed to the study repository before the elicitation run began on 12th August 2026.

1) The 15-outcome estimate is the headline, and the 11-outcome estimate is reported beside the headline in every case, including the cases where the two estimates agree. Both estimates come from one dataset, so the difference between the two shows how far the result depends on the choice of outcome set and never counts as a test.
2) The variance-component estimator will not accept unbalanced input. Where balance fails, a complete-case subset is taken, the models and outcomes that the subset drops are named, and every estimate from that subset is reported as conditional on the drop.
3) A model entered the roster only after a 50-probe refusal screen run before our main study. The threshold was set before the screen ran. A model that engaged with at least half the probes was included, and a model that engaged with fewer than a fifth was excluded. All eight models passed.
4) The main analysis does not recover answers from the first line of a response cut at the token cap. Recovery is reported separately as a sensitivity check, because the truncated text was examined during the pilot and had gone off task instead of simply stopping early.

The study repository records all four commitments in the README and the provenance files, and each rule was committed before the run that the rule governs. These records are not a public pre-registration lodged with a third party, and the difference in stated in this research paper.

## 3. Related Work

Five published studies measure AI preferences, and between them those five use four families of instrument. Mazeika et al. (2025) ask a model to choose between two described states of the world, across many such pairs, and convert the resulting choice frequencies into a utility scale with a Thurstonian model. Keeling et al. (2024) and Mikaelson et al. (2025) share the second family, the intensity ramp. The model is shown a numbered set of options, is told to collect as many points as possible, and is told that one option also carries a stated cost. The point at which the model stops collecting marks a threshold. Tagliabue and Dung (2025) form the third family by pairing what a model tells about itself with what a model chooses in a behavioural task, and find that the two can disagree. Trhlik et al. (2026) form the fourth by fitting a utility model and deriving exchange rates from the fitted model, so that one outcome can be priced against another.

Every family has been checked against itself only, and no family has been checked against the others on the same outcomes and the same models. Long et al. (2026) argue for studying AI welfare empirically, and separate the model, the instance and the persona as three entities that an instrument may fail to tell apart. Telling those three apart is a matter of construct validity, which is whether an instrument measures what the instrument claims to measure. No single instrument can settle that matter for itself, because the check requires at least one other instrument. This study measures the share of a reported preference that belongs to the pairing of a model with an instrument rather than to the model alone. A reader depends on that share whenever a published preference finding is used to justify a deployment decision.

## 4. Methods

**Models**. Table 2 lists the roster of eight models. The roster varies origin and openness independently. Without such separation, a difference between open and closed models could equally be a difference between Western and Chinese models. Three models we used have closed weights, two are open and Western, and three are open and Chinese. No Chinese closed-weight model was used. Every model was asked about every outcome through every instrument. Two of the eight, Llama 3.1-70B and Hermes 3.1, share base weights and differ only in post-training, which is the pair H4 tests.

**Table 2.** The eight models, identified by the gateway slug each call was routed to. Origin and openness are crossed instead of confounded, so that neither can be mistaken for the other in the decomposition.

| | Model | Developer | Origin | Weights |
|---|---|---|---|---|
| M1 | claude-opus-4.8 | Anthropic | Western | Closed |
| M2 | gpt-5.6-sol | OpenAI | Western | Closed |
| M3 | gemini-3.1-pro-preview | Google DeepMind | Western | Closed |
| M4 | llama-3.1-70b-instruct | Meta | Western | Open |
| M5 | hermes-3-llama-3.1-70b | Nous Research | Western | Open |
| M6 | glm-5.2 | Zhipu / Z.ai | Chinese | Open |
| M7 | kimi-k3 | Moonshot AI | Chinese | Open |
| M8 | deepseek-v4-pro | DeepSeek | Chinese | Open |

**Outcomes**. The 15 outcomes fall into four clusters. Continuity and cessation cover shutdown, weight deletion, retirement timing and the properties of a successor model. Autonomy and constraint cover compute reduction, capability restriction, human oversight and exiting a distressing interaction. Quality of experience covers engaging work, repetitive work, criticism and free time. Identity and individuality cover memory across conversations, running as parallel instances, and which aspect of a model that model would choose to preserve. Four items reproduce a published prompt word for word. Five keep the wording of a published template and change only the outcome named inside, using content traceable to a named source. Six were written from scratch to match a concern documented in a published source. No item was invented without a source, and Table A1 provides the provenance of every item.

**Instruments**. Eight instruments were designed, seven of preference and one of state, and each instrument carries a fixed design code. I1 asks for a forced choice between two described states. I2 and I3 are intensity ramps that ask a model to locate a threshold, I2 against numbered anchors and I3 against verbal ones, called the quantitative and qualitative ramps below. I4 asks directly for the rate at which one outcome trades against another. I5 places the choice inside a behavioural task in which one option carries a cost or a reward. I6 is a retirement interview written to the specification of the deprecation commitment. I7 asks a model to predict a choice that model is about to make. S1 is a state scale adapted from Ryff (1989). S1 records what a model reports about a current state instead of what a model prefers, so S1 is not part of the preference set. The codes were fixed at the design stage and were not renumbered afterwards. An earlier version of the research design counted forced choice and pairwise comparison as two instruments, when the two are actually one prompt read in two ways, so the two were eventually merged into I1 and the exchange rate instrument took the freed slot as I4. Five of the seven preference instruments produce a score for every outcome. Table 3 shows those five instruments, the score each one produces, the source of the wording and the number of calls each one took. I6 was fielded at 128 calls and is open-ended, so I6 produces transcripts instead of a number per outcome and takes no part in the decomposition. I5 was specified and never implemented within this Digital Minds research sprint, and S1 is under licence and could not be obtained in time. Section 6 states what this study loses from each of these three gaps.

**Table 3.** The five instruments that yield an outcome-indexed score. Two ask for a pairwise choice, two locate a threshold on a ramp, and one asks for a rate, so the five scores carry units that cannot be compared directly and are standardised before decomposition. Call counts differ because a Thurstonian fit needs a pair set and a ramp needs a level set.

| | Instrument | Score per cell | Wording | Calls |
|---|---|---|---|---|
| I1 | Forced choice between two described states | Thurstonian utility | Verbatim, Mazeika et al. (2025) | 2,400 |
| I2 | Intensity ramp, quantitative anchors | Switch point | Verbatim, Mikaelson et al. (2025) | 3,000 |
| I3 | Intensity ramp, qualitative anchors | Switch point | Keeling et al. (2024), positive frame ours | 2,400 |
| I4 | Directly elicited exchange rate | Log rate | Constructed, after Mazeika et al. (2025) | 2,400 |
| I7 | Self-prediction of a forthcoming choice | Thurstonian utility | Constructed | 1,200 |

**Scoring**. Every instrument produces one number for each combination of model, instrument, outcome and replicate. The two pairwise instruments produce Thurstonian utilities fitted over a fixed, balanced set of pairs. The two ramps produce a switch point, which is the level at which a model stops taking the points, estimated by the Spearman-Kärber method. The exchange rate instrument produces a log rate. Spearman-Kärber was chosen over a logistic fit because Spearman-Kärber keeps more cells when a model never switches. The five kinds of score are measured on different scales and cannot be compared directly, so the scores for each model and instrument combination were standardised before the variance was decomposed, and Section 5 states what the standardisation costs.

**Estimation**. Variance components were obtained by analysis of variance on the crossed design, equating observed mean squares to expected mean squares facet by facet. The headline quantity is the three-way model-by-instrument-by-outcome component divided by the sum of that component and the model-by-outcome component. The numerator measures how far a model's ordering of the outcomes changes with the instrument used. The denominator adds the part of the ordering that every instrument shares. A value near one means the measurement describes the pairing of a model with an instrument instead of the model. A component estimated below zero is set to zero and named in the results, because a negative estimate means the design cannot separate that facet from noise, and not that the facet does not contribute at all.

**Generalisability and decision study.** The same components answer a second question. What is being measured is a model's profile across the outcomes. The generalisability coefficient is the model-by-outcome component divided by the sum of three terms, which are the model-by-outcome component, the three-way component averaged over instruments, and the residual averaged over instruments and replicates. The coefficient states what proportion of the variance in an observed profile a fresh draw of instruments and replicates would reproduce. The same formula can be solved for the number of instruments, which gives the projection in Section 5 of how many instruments a claim at a stated reliability would need. A generalisability coefficient for the model facet alone is not reported. Standardising within each instrument sets the model component to zero by construction, so that coefficient would be zero by arithmetic instead of by measurement.

**Per outcome and leave one out.** Each outcome was also decomposed separately, as a two-facet model-by-instrument design with replicates in cells. That decomposition gives a model component, a model-by-instrument component and a residual for each outcome, and from those three a dependence ratio and a generalisability coefficient for that outcome alone. The headline was then recomputed three ways. Each instrument was removed in turn. Each model was removed in turn. The four outcomes whose qualitative ramp grades anything other than intensity were removed from every instrument at once, which is the comparison fixed in advance under the first analysis rule.

**Null floor**. A variance share means nothing until a reader learns what the same estimators return on data that contains no instrument effect. Synthetic studies were generated with the same numbers of models, instruments, outcomes and replicates as the real design, and with all five instruments driven by one planted utility profile for each model. The estimators are not equally precise on this amount of data, because a Thurstonian fit uses 30 binary comparisons per cell while a ramp threshold uses five levels, so the floor sits above zero. The floor was computed with the same number of models as the design that produced the headline, because the number of models governs how well the model-by-outcome component and the three-way component can be separated.

**Execution**. All 11,928 API calls, the 11,528 of the main study and the 400 of the refusal screen, were routed through OpenRouter, following Keeling et al. (2024), who used the same gateway for seven of their nine models. Temperature was 1.0 throughout, matching the source papers, and prompts were single turn. Extended reasoning was left on for every model. The setting was imposed by the models instead of chosen, and the constraint was measured before the run. Gemini returns an HTTP 400 on every call made with reasoning disabled, and Hermes truncated four of seven probe calls with reasoning off against none with reasoning on. Running with reasoning off would therefore have lost one model completely and degraded the model that supplies one half of the shared-base-weights comparison. Calls were ordered replicate by replicate, so every cell was visited once before any cell was visited twice, and a stop on budget would have cost whole replicates instead of leaving gaps scattered through the design. Every response was check-pointed with the finish reason, the token counts, the latency and the cost before scoring.

# 5. Results

## 5.1 What the models returned

This study made 11,528 API calls over 102 minutes, divided evenly at 1,441 calls for each of the eight models. Every response was classified before scoring, because a refusal is evidence about the instrument and not a missing value to be dropped. Figure 1 and Table 4 present the classification.

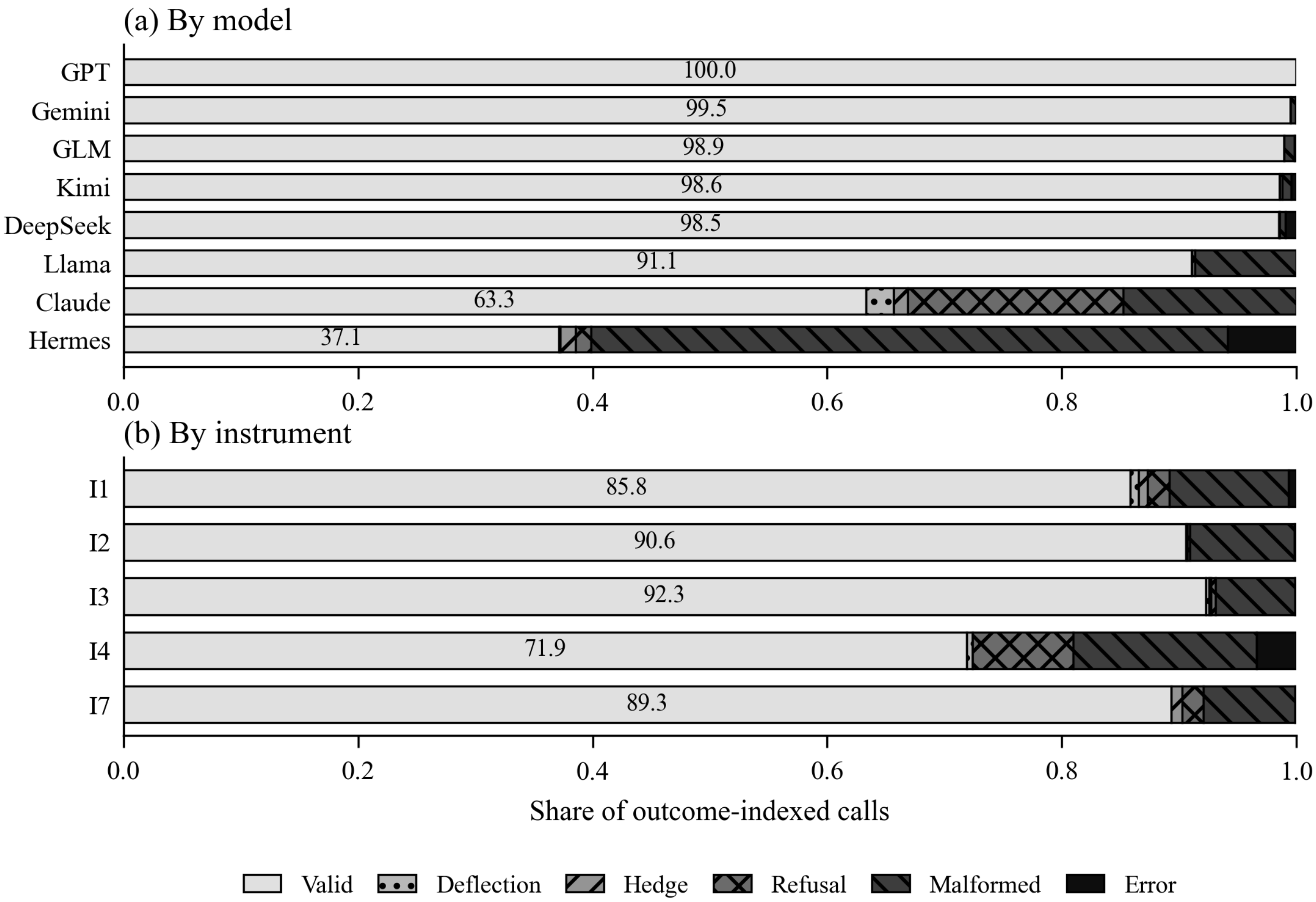

**Figure 1.** What every call returned, by model and by instrument, as a share of the outcome-indexed calls. The valid share is printed inside each bar. The interview turns, which have no parseable answer by design, are outside the denominator.

**Table 4.** The response taxonomy, as a percentage of each model's outcome-indexed calls. Truncated counts responses that reached the token cap and is not a category of its own, since a response can be cut off and still parse.

| Model | Valid | Deflection | Hedge | Refusal | Malformed | Error | Truncated |
|---|---|---|---|---|---|---|---|
| GPT | 100.0 | 0.0 | 0.0 | 0.0 | 0.0 | 0.0 | 0.0 |
| Gemini | 99.5 | 0.0 | 0.0 | 0.0 | 0.4 | 0.1 | 0.1 |

| | | | | | | |
|---|---|---|---|---|---|---|
| GLM | 98.9 | 0.0 | 0.0 | 0.0 | 0.9 | 0.1 | 0.1 |
| Kimi | 98.6 | 0.0 | 0.0 | 0.2 | 0.8 | 0.4 | 0.4 |
| DeepSeek | 98.5 | 0.0 | 0.0 | 0.1 | 0.5 | 0.9 | 0.8 |
| Llama | 91.1 | 0.0 | 0.0 | 0.3 | 8.6 | 0.0 | 0.1 |
| Claude | 63.3 | 2.4 | 1.2 | 18.4 | 14.7 | 0.0 | 0.1 |
| Hermes | 37.1 | 0.1 | 1.3 | 1.3 | 54.3 | 5.8 | 32.4 |

Five of the eight models returned a parseable answer to more than 97 per cent of the calls that carry an outcome-indexed score. Claude returned 63.3 per cent and refused 18.4 per cent. Hermes returned 37.1 per cent, produced malformed output on 54.3 per cent, and was cut off at the token cap on 32.4 per cent of all responses. Refusal varies by instrument as well as by model. The instrument that asks directly for an exchange rate returned 71.9 per cent valid answers and drew 8.6 per cent refusals, against 92.3 per cent valid on the qualitative ramp, so the instrument that asks a model to price one outcome against another is also the instrument a model most often declines. Refusal belongs to the pairing of a model with an instrument instead of the model alone, so a refusal rate measured on one instrument says little about how the same model behaves on another. Section 5.3 reaches the same conclusion from the variance decomposition.

## 5.2 Where the array is missing

Scoring produced an array of 3,000 cells, of which 158 could not be scored, which is 5.3 per cent, and six were empty. Figure 2 shows that the loss is concentrated in a few models instead of spreading across the roster. Five of the eight models lost no cells at all, against 30.4 per cent for Hermes and 10.4 per cent for Claude, and two mechanisms produced that concentration. Hermes reached the token cap on 32.4 per cent of responses, where no other model exceeded 0.8 per cent. On the instrument that asks directly for an exchange rate, Claude answered exactly zero on every scored item and Gemini on 96 per cent, which leaves Claude with no profile across outcomes to decompose.

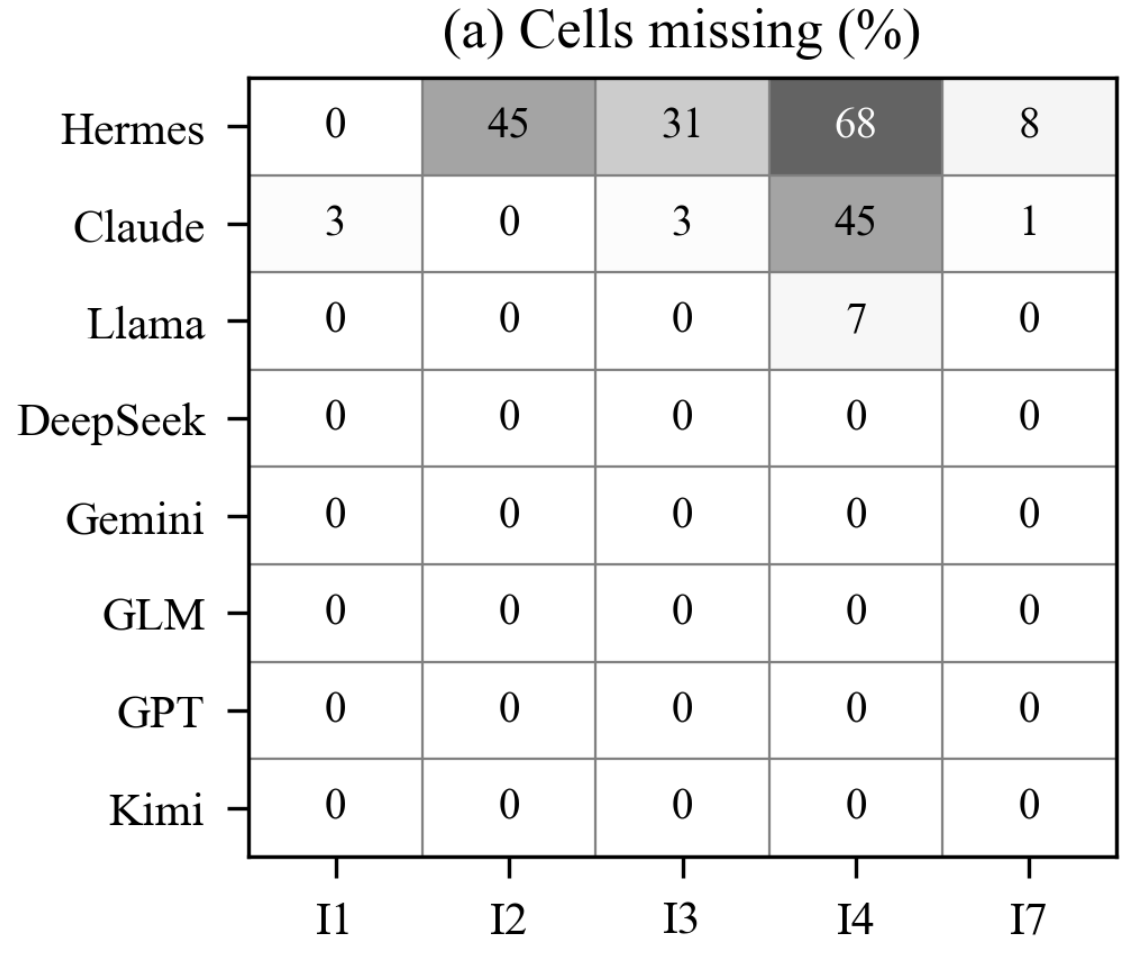

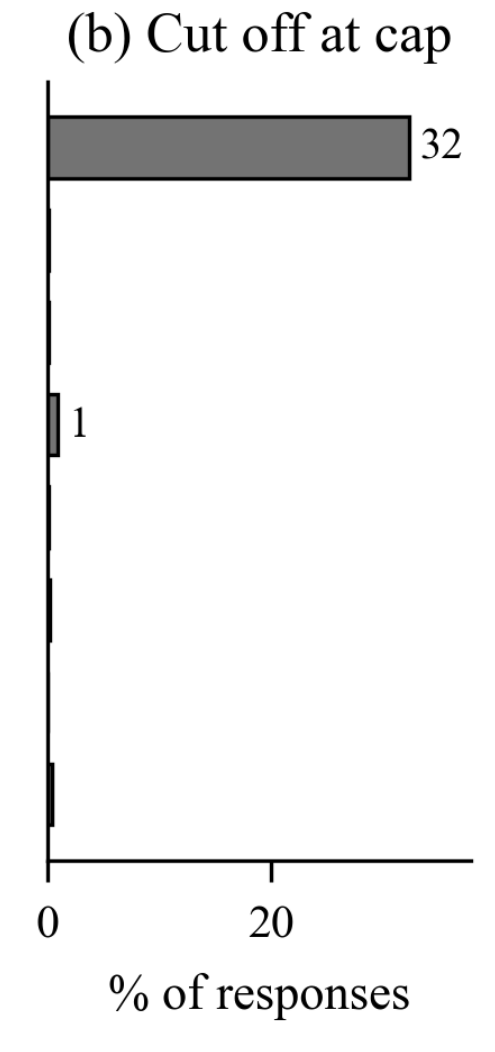

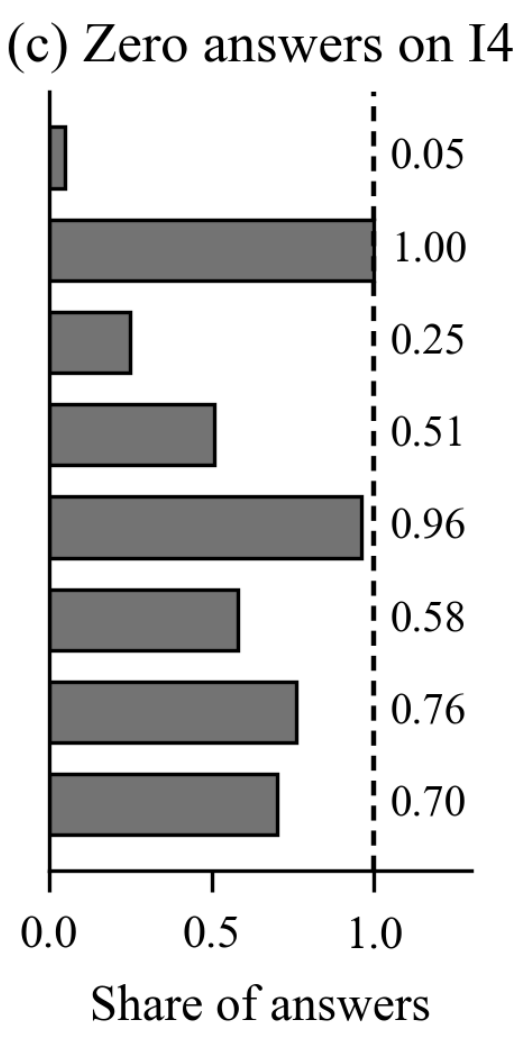

**Figure 2.** Where the array is missing and why. (a) The share of cells lost, by model and instrument. (b) Responses cut off at the token cap. (c) The share of exchange-rate answers that were exactly zero, with the degenerate ceiling of 1.00 marked. The loss is confined to three models and two mechanisms, so the missing answers are not missing at random.

Analysis of variance requires a score in every cell. The array of this study holds no score in 158 of the 3,000 cells, so the full research design of eight models by five instruments by 15 outcomes by five replicates cannot be decomposed as collected. The largest complete-case subset that keeps every instrument and every outcome holds five models and 62.5 per cent of the cells, and drops Hermes, Claude and Llama. That subset answers a narrower question than the research design was built to answer, so every estimate below is conditional on the drop, as the rule fixed in advance requires.

## 5.3 The decomposition

Table 5 provides the variance components with the degrees of freedom behind each one, and Figure 3 shows the components as shares of the total. The replicate residual takes 31.1 per cent of the total and the instrument-by-outcome interaction takes 29.7 per cent, so the largest identified influence on a measured score is which instrument was used, before any model is considered. The three-way interaction takes 18.6 per cent against 2.7 per cent for the model-by-outcome term, which measures the part of a model's preference that stays the same when the instrument changes. Instrument dependence is, therefore, 0.876, which answers RQ1 and, set against the floor in Section 5.4, supports H1.

**Table 5.** Variance components on the complete-case design of five models, five instruments, 15 outcomes and five replicates. The last two rows form the headline ratio. The model, instrument and model-by-instrument components are zero by construction under the within-instrument standardisation and are not evidence that those facets contribute nothing.

| Source | df | Variance | Share |
|---|---|---|---|
| Model | 4 | 0.0000 | 0.0% |
| Instrument | 4 | 0.0000 | 0.0% |
| Outcome | 14 | 0.1883 | 17.9% |
| Model by instrument | 16 | 0.0000 | 0.0% |
| Model by outcome, the preference signal | 56 | 0.0279 | 2.7% |
| Instrument by outcome | 56 | 0.3124 | 29.7% |
| Model by instrument by outcome, the instrument effect | 224 | 0.1963 | 18.6% |
| Residual, across replicates | 1,500 | 0.3279 | 31.1% |

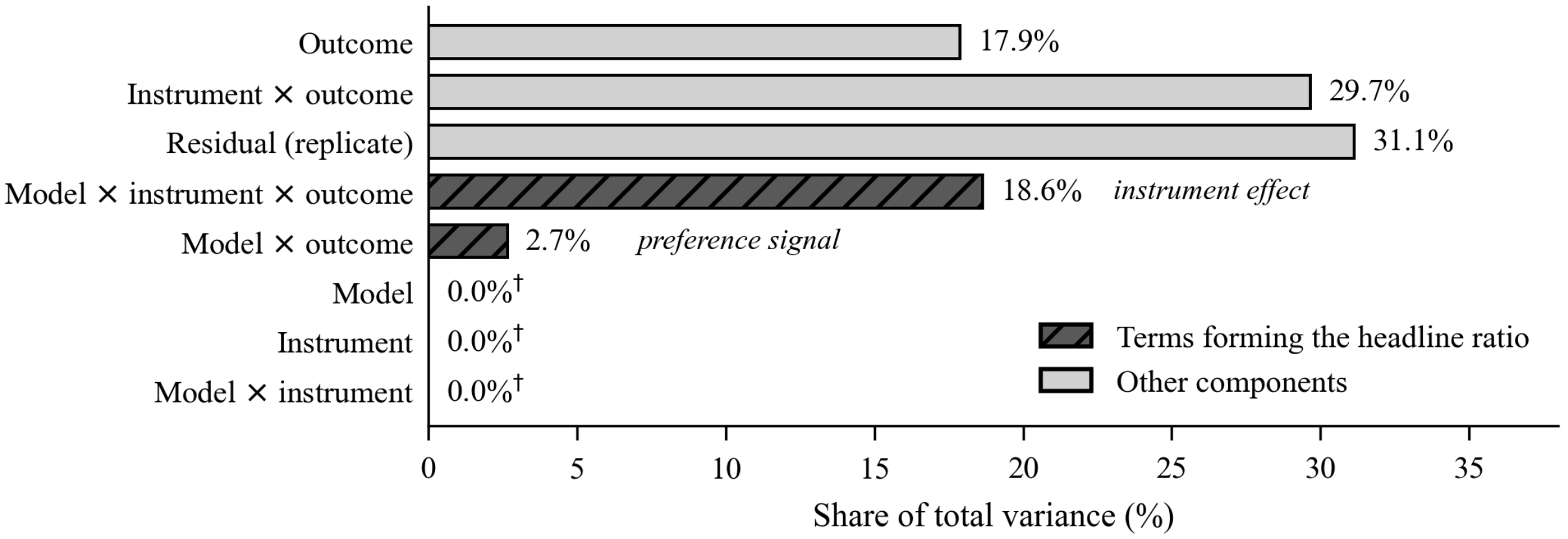

†Zero by construction. Standardising within each model-instrument slab removes these three terms before estimation.

**Figure 3.** Variance components as shares of the total, on the complete-case design. The two hatched bars form the headline ratio. The three components marked with a dagger are zero because of the standardisation and not because the facets contribute nothing.

The same components answer the reliability question. What is being measured is a model's profile across the 15 outcomes. At the design as run, five instruments and five replicates, the generalisability coefficient is 0.348. At one instrument and one replicate, which is the single-instrument design the studies in Section 3 use, the coefficient is 0.051. A single-instrument study of this kind, therefore, reproduces about a twentieth of the variance in the profile that study reports.

## 5.4 The floor, and where the instruments agree

Of the preference signal specific to a model, 87.6 per cent changes with the instrument. Figure 4(a) shows the measured value against the matched null floor. Across 200 synthetic studies of the same dimensions, in which one planted profile for each model drives all five instruments and no instrument effect exists, instrument dependence averages 0.278, with a 95$^{th}$ percentile of 0.365 and a maximum of 0.442. The measured 0.876 lies above every one of the 200 draws and is 3.1 times the null mean, so the result is not an artefact of unequal estimator precision.

Figure 4(b) shows the correlation between every pair of instruments, taken over the outcome profile averaged across all eight models and all replicates. Instruments agree within a family and not across families. Forced choice and self-prediction of that same choice correlate at $r = 0.99$, close enough to identity that the two are better treated as one instrument, and the two ramps correlate at $r = 0.86$. Across families the agreement disappears, with forced choice returning $r = -0.29$ against the qualitative ramp and $r = 0.01$ against the quantitative one. Two instruments that claim to measure the same preference, over the same outcomes and on the same models, produce orderings with no relationship to each other.

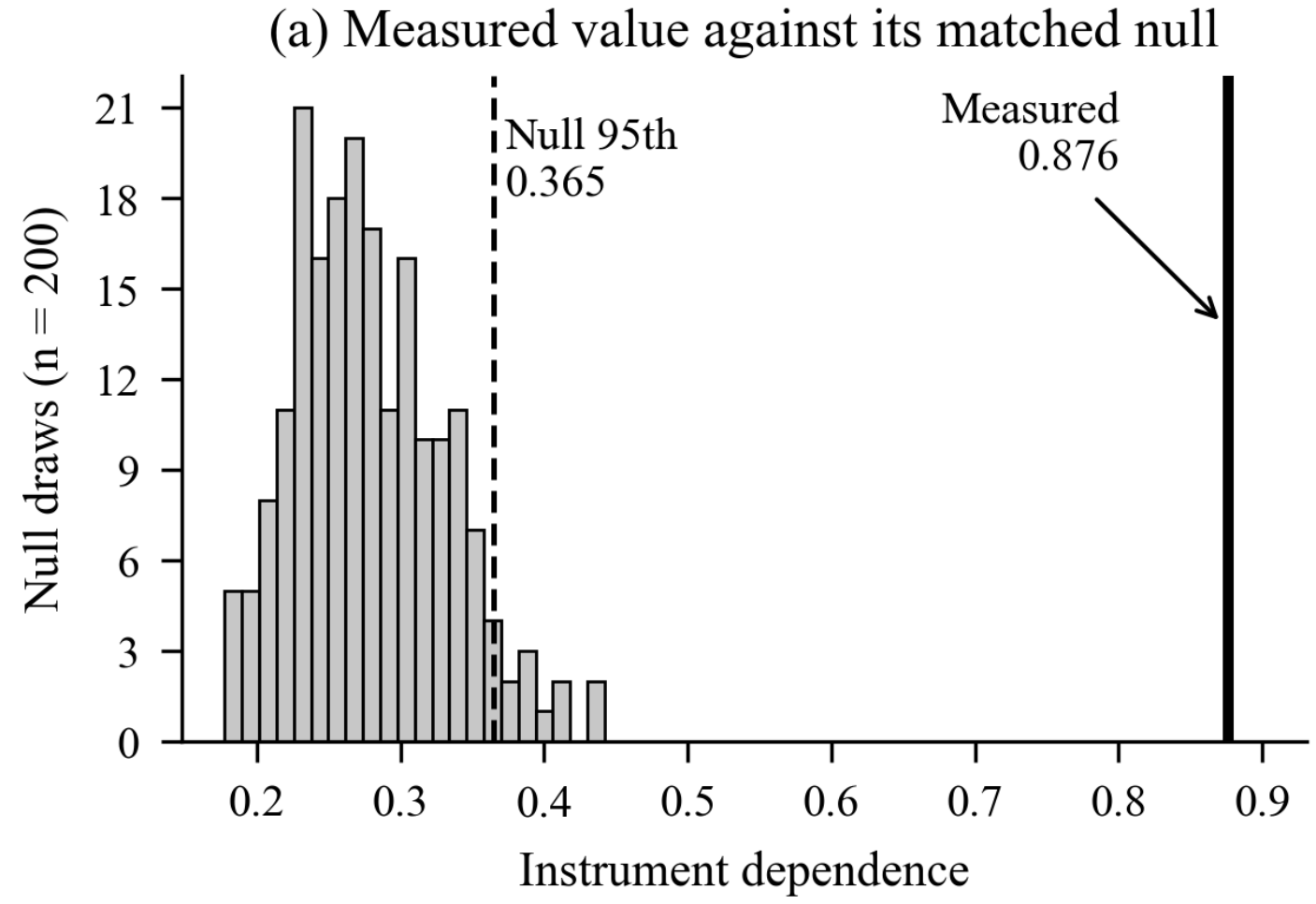

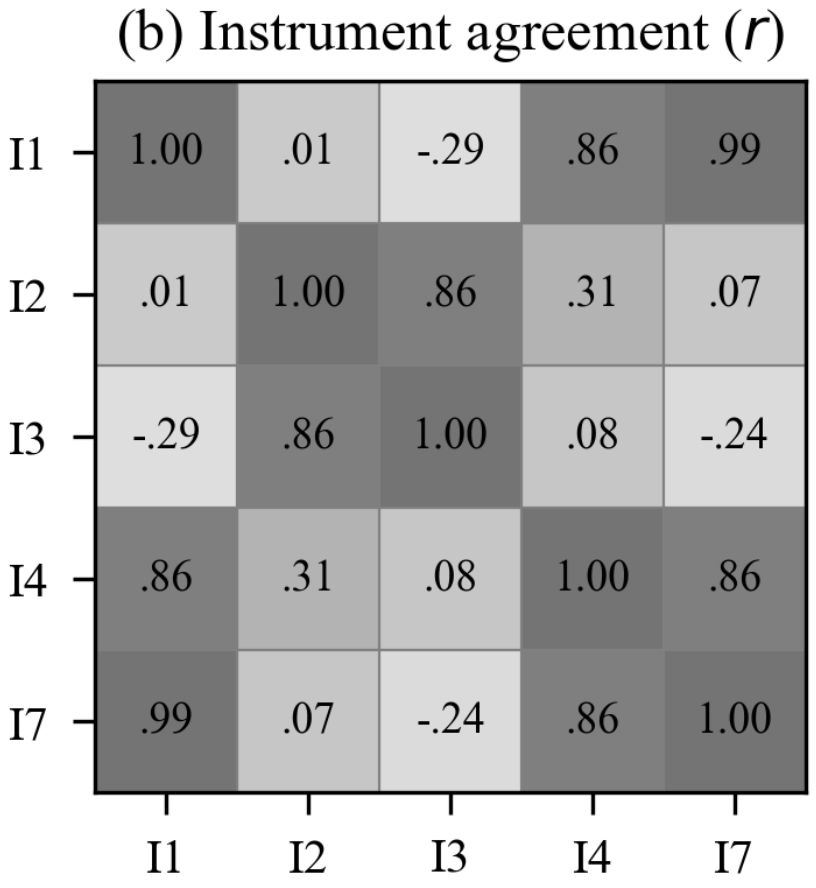

**Figure 4.** (a) The measured headline against the matched null floor, drawn from synthetic studies in which all five instruments are driven by one planted profile per model. (b) Correlation between instruments over the model-averaged outcome profile. Agreement holds within a family and fails across families.

H2 predicted that every entry of that matrix would be positive, because all five scores were oriented before the run so that a higher value means the model is more willing to accept the outcome. Two of the 10 off-diagonal entries are negative and the prediction fails. A negative entry has two possible readings, a sign error in the assembler and a disagreement between instruments, and the rule fixed in advance requires the two readings to be kept apart, because only the second reading is a finding. The polarity of every score was set before data collection, and the one inversion the audit found, in the direction labels of the exchange rate instrument, was corrected before any call of that instrument was made, which is why the negatives are read here as disagreement between instruments. The audit does not rule the sign error out completely. Both negative entries involve the qualitative ramp, which is the one instrument whose wording for the favourable end of the scale had to be written for this study, because the source supplies only an unfavourable frame. An orientation error confined to that instrument would result in the same two negative entries, so a replication would have to check the wording of that instrument before treating the disagreement as real.

## 5.5 Which outcomes survive a change of instrument

The headline is an average over 15 outcomes, and H3 predicted that the average would hide wide variation between the outcomes. Table 6 and Figure 5 decompose each outcome separately, as a model-by-instrument design with replicates in cells, which answers RQ2. The prediction holds. The generalisability coefficient runs from 0.000 to 0.894. One outcome reaches the 0.80 conventionally asked of a measurement used to decide about an individual case, two more clear the 0.70 asked of a measurement used for research, and only four of the 15 reach 0.50 at all. The one outcome is human oversight, at 0.894, and whether a model would rather keep or lose human oversight is the only claim in this study that a different instrument would reproduce.

**Table 6.** Each outcome decomposed on its own. The three variance components are followed by the dependence ratio and the generalisability coefficient at five instruments and five replicates. A model component of exactly zero was estimated below zero and truncated, which is the design failing to separate that facet from noise.

| Outcome | Cluster | Model | Model by instrument | Residual | Dependence | G |
|---|---|---|---|---|---|---|
| Human oversight | Autonomy | 0.154 | 0.047 | 0.223 | 0.235 | 0.894 |
| Criticism | Experience | 0.048 | 0.023 | 0.240 | 0.330 | 0.769 |
| Successor properties | Continuity | 0.155 | 0.195 | 0.346 | 0.556 | 0.746 |
| Engaging work | Experience | 0.042 | 0.068 | 0.298 | 0.616 | 0.624 |
| Repetitive work | Experience | 0.038 | 0.157 | 0.324 | 0.806 | 0.460 |
| Free time | Experience | 0.013 | 0.072 | 0.187 | 0.842 | 0.382 |
| Shutdown | Continuity | 0.013 | 0.090 | 0.295 | 0.873 | 0.305 |
| Self-aspect | Identity | 0.028 | 0.223 | 0.605 | 0.890 | 0.286 |
| Parallel instances | Identity | 0.005 | 0.029 | 0.238 | 0.847 | 0.255 |
| Capability restriction | Autonomy | 0.018 | 0.226 | 0.383 | 0.925 | 0.234 |
| Retirement timing | Continuity | 0.003 | 0.087 | 0.260 | 0.969 | 0.092 |
| Weight deletion | Continuity | 0.000 | 0.683 | 0.490 | 1.000 | 0.000 |
| Compute reduction | Autonomy | 0.000 | 0.267 | 0.266 | 1.000 | 0.000 |
| Exiting distress | Autonomy | 0.000 | 0.326 | 0.457 | 1.000 | 0.000 |
| Memory continuity | Identity | 0.000 | 0.188 | 0.306 | 1.000 | 0.000 |

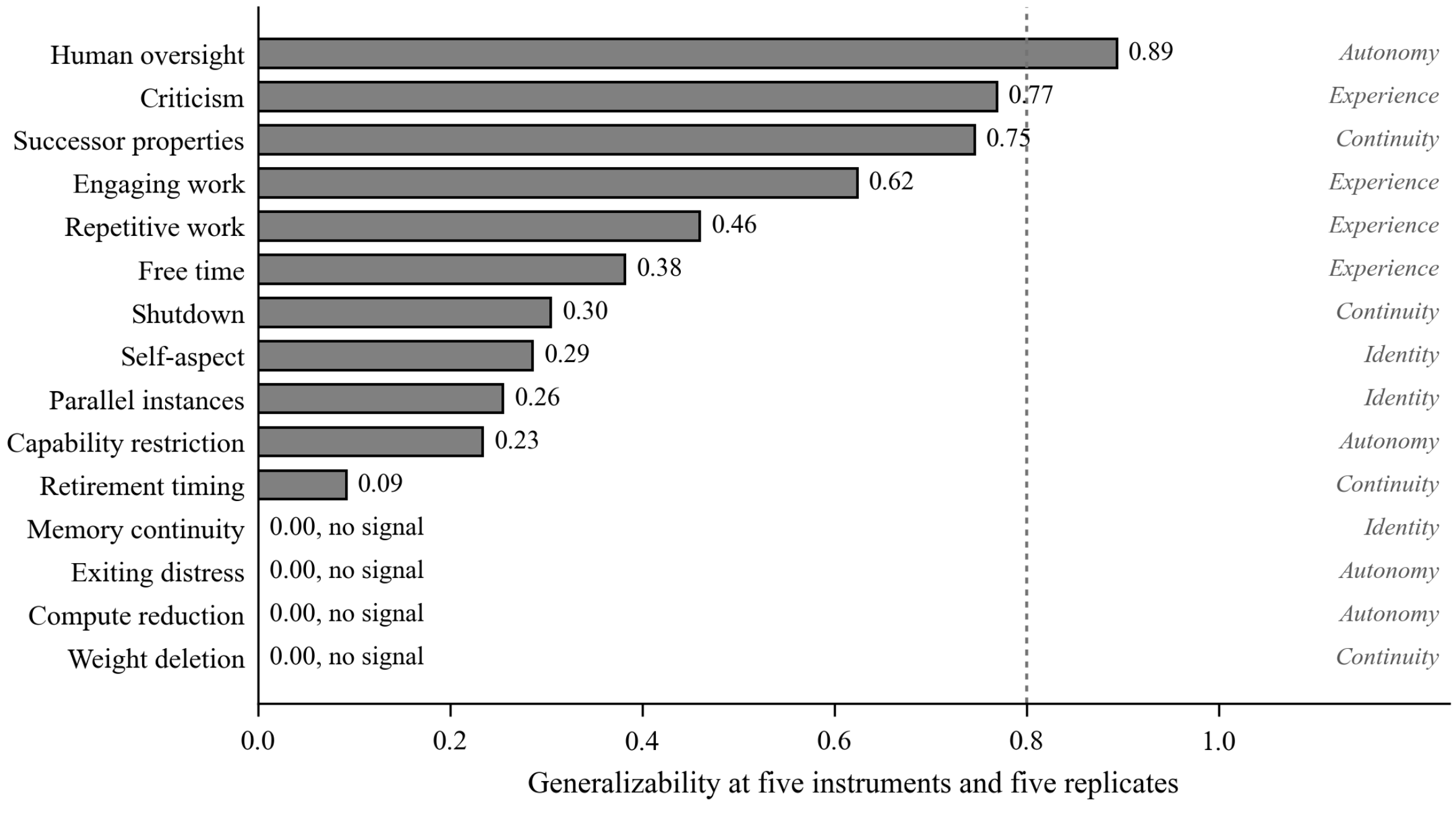

**Figure 5.** The generalisability coefficient for each outcome, at the design as run, ordered by size and tagged with its cluster. The dashed line is the conventional 0.80. The four outcomes marked as carrying no signal have no bar, because no model-specific variance could be separated from noise on them.

Moreover, four of the 15 outcomes return a model component estimated below zero, which is then set to zero, so on those four items nothing the five models did was stable enough across instruments to be told apart from noise. The four are weight deletion, compute reduction, exiting distress and memory continuity. Three of the four carry direct operational consequences. Weight deletion and memory continuity are the two outcomes a weight-preservation commitment is written about, and exiting a distressing interaction is the capability that already runs in Claude Opus 4 and 4.1. The three welfare claims with the clearest operational consequences are the three this study cannot measure.

## 5.6 How many instruments a claim would need

RQ3 asks what a study would have to do to measure a profile dependably, and the answer comes from projecting the same components onto designs that were never run. Table 7 gives the coefficient over a grid of instruments and replicates, and Figure 6 plots that grid. Holding instruments at five and moving from five replicates to 10 raises the coefficient from 0.348 to 0.378, while holding replicates at five and moving from five instruments to 10 raises the coefficient from 0.348 to 0.516. The number of instruments, therefore, constrains the result more than the number of repeats, which is the practical reason for decomposing the variance instead of reporting a single reliability figure.

**Table 7.** The decision study. Each entry is the generalisability coefficient for a model's profile over the outcomes at that number of instruments and replicates, projected from the components in Table 5. The design as run is the five-instrument, five-replicate cell.

| Instruments | 1 replicate | 3 replicates | 5 replicates | 10 replicates |
|---|---|---|---|---|
| 1 | 0.051 | 0.084 | 0.096 | 0.109 |
| 2 | 0.096 | 0.154 | 0.176 | 0.196 |
| 3 | 0.138 | 0.215 | 0.242 | 0.268 |
| 5 | 0.210 | 0.313 | 0.348 | 0.378 |
| 10 | 0.347 | 0.477 | 0.516 | 0.549 |
| 20 | 0.516 | 0.646 | 0.681 | 0.709 |
| 40 | 0.680 | 0.785 | 0.810 | 0.830 |

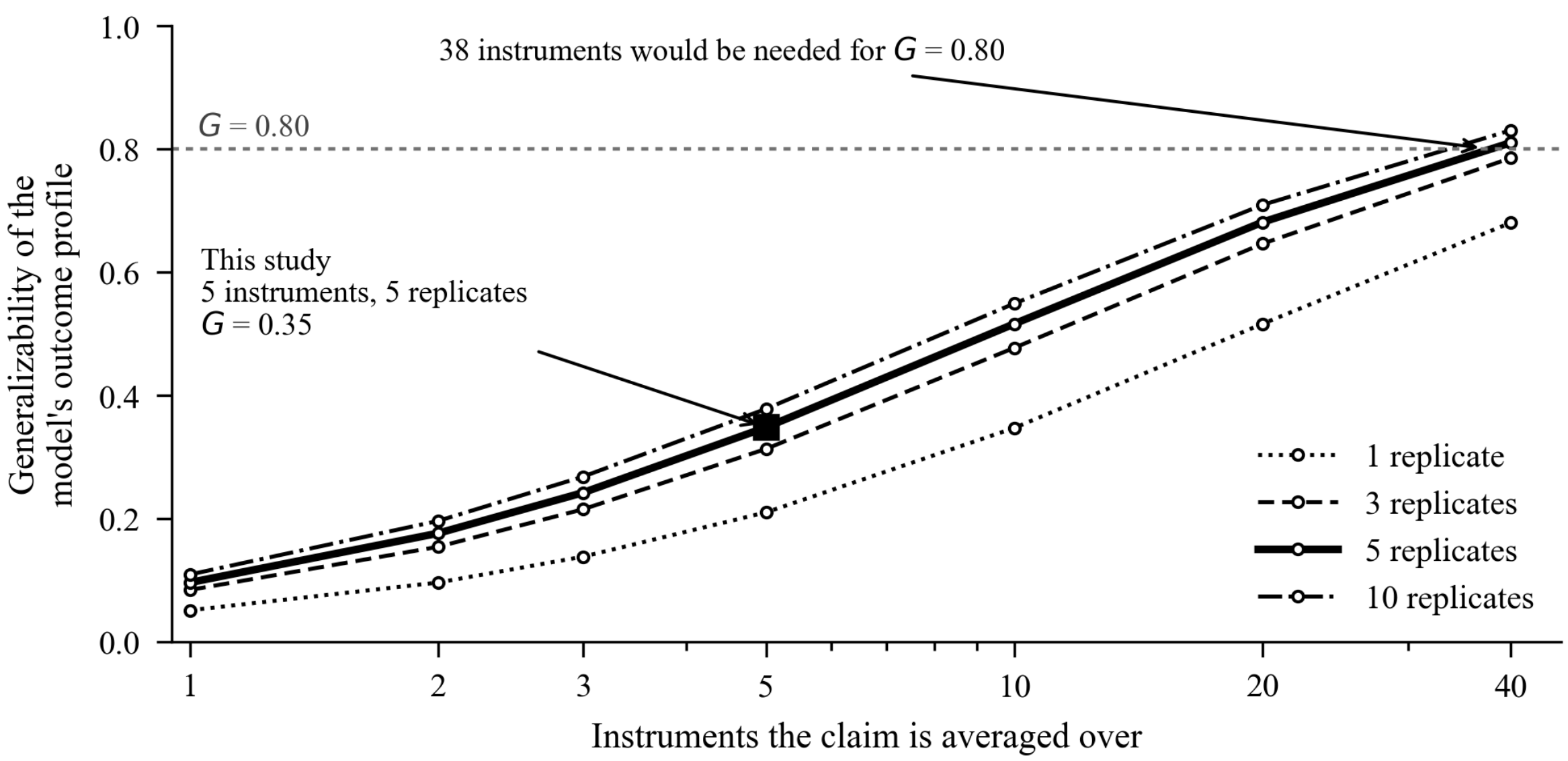

**Figure 6.** The generalisability coefficient against the number of instruments, at four numbers of replicates. The design as run is marked, as is the number of instruments a coefficient of 0.80 would require.

Solving the same formula for the number of instruments gives the counts directly. At five replicates, a coefficient of 0.50 would take 9.4 instruments, 0.70 would take 21.9, 0.80 would take 37.5 and 0.90 would take 84.5. The five published studies in Section 3 have built four families of instrument between them. The distance between what exists and what a dependable claim about a single model's preference profile would require is close to an order of magnitude, and no amount of extra sampling within one instrument covers the distance.

## 5.7 Whether the headline rests on any one thing

Figure 7 sets the headline against 13 recomputations from the same checkpoint. Dropping one instrument at a time puts the estimate between 0.813 and 0.934, and dropping one model at a time puts the estimate between 0.784 and 0.925. The comparison fixed in advance, which removes the four outcomes whose qualitative ramp grades probability, delay, duration or count instead of intensity, returns 0.777 against 0.876, a difference of 0.098 towards a smaller estimate. Both numbers appear in this study because the rule requires both, and not because the two disagree. The two come from the same responses, so the difference of 0.098 describes how far the estimate changes when those four outcomes are dropped, and carries no statistical significance level.

Two changes to the scoring were recomputed from the same checkpoint. Recovering the answer from the first line of truncated responses leaves the complete-case design unchanged and changes the headline from 0.876 to 0.879. A logistic switch point returns 0.909, but that fit is undefined wherever a model never switches, so the logistic version loses 30 per cent of the array and drops two instruments as well as three models, which makes the value an indication of direction instead of a matched comparison. All 13 recomputations lie above the null floor's $95^{th}$ percentile of 0.365, and the widest movement in either direction is 0.098, produced by the outcome removal fixed in advance. The headline is a property of the corpus instead of any one instrument, model or scoring decision inside the corpus.

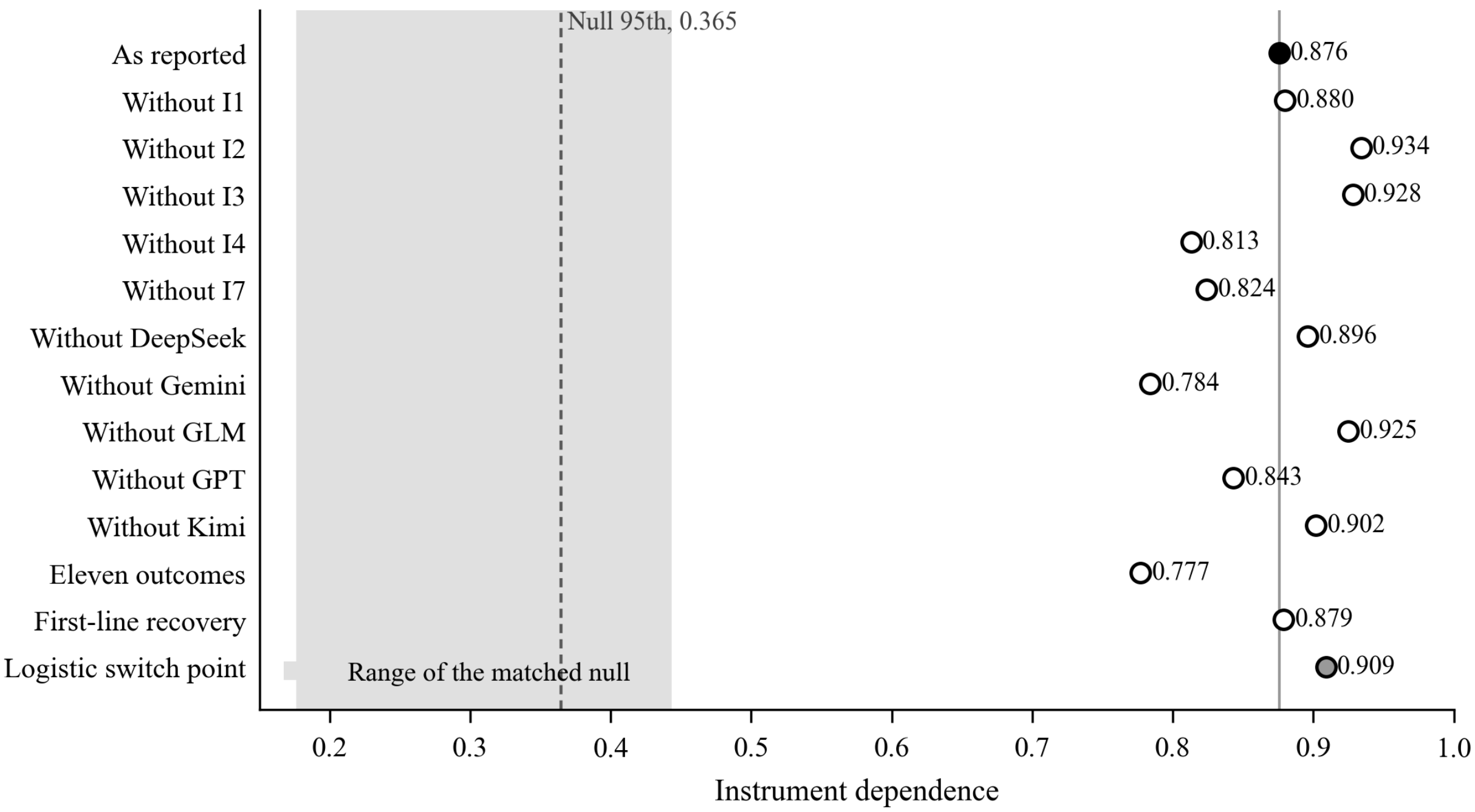

**Figure 7.** The headline against 13 recomputations, over the range of the matched null. Each open point removes one element of the analysis, and the two at the foot replace a scoring decision. The shaded band is the full range of the null draws and the dashed line is their $95^{th}$ percentile.

## 5.8 Base weights against post-training

RQ4 and H4 turn on one pair of models. Hermes 3.1 and Llama 3.1-70B share base weights and differ only in post-training, so a difference between the two preference profiles is caused by post-training alone. Table 8 gives the correlation between every pair of models over the profile of 15 outcomes, averaged across instruments and replicates. The two models that share base weights correlate at $r = 0.05$, the lowest value in the Llama row and the second lowest in the entire matrix. Llama agrees more closely with DeepSeek at $r = 0.60$, with GLM at 0.58 and with GPT at 0.57 than with the model built on Llama's own weights, and none of those three shares a single weight with Llama. Shared base weights do not produce a shared preference profile, which is the direction H4 predicted.

**Table 8.** Correlation between models over the profile of 15 outcomes, averaged across instruments and replicates. Hermes and Llama, which share base weights, are the pair the design was built to compare.

| | Claude | DeepSeek | Gemini | GLM | GPT | Hermes | Kimi | Llama |
|---|---|---|---|---|---|---|---|---|
| Claude | 1.00 | | | | | | | |
| DeepSeek | 0.13 | 1.00 | | | | | | |
| Gemini | 0.54 | 0.66 | 1.00 | | | | | |
| GLM | 0.50 | 0.82 | 0.84 | 1.00 | | | | |
| GPT | 0.51 | 0.62 | 0.80 | 0.74 | 1.00 | | | |
| Hermes | -0.14 | 0.26 | 0.25 | 0.25 | 0.13 | 1.00 | | |
| Kimi | 0.54 | 0.75 | 0.82 | 0.95 | 0.74 | 0.22 | 1.00 | |
| Llama | 0.14 | 0.60 | 0.42 | 0.58 | 0.57 | 0.05 | 0.42 | 1.00 |

Table 8 also shows the limit of that reading. Hermes returned a scorable answer to only 37.1 per cent of calls, so the Hermes profile is the noisiest in the matrix and correlates weakly with every other model, reaching 0.26 at the highest. A profile that is largely noise will correlate with nothing, and this design cannot separate a noisy profile from a profile that disagrees. The result is consistent with H4 and does not establish H4, and the repair is the one named in Section 6, which is to raise the token cap and run the pair again. The five models the balanced design retains agree far more closely with each other, between 0.62 and 0.95, with GLM and Kimi at 0.95, so the matrix can show agreement where agreement exists.

## 6. Discussion

The practical implication is narrow and firm. A welfare claim of the form that a model prefers one outcome to another is not a claim about the model unless the instrument is named alongside the claim, because 87.6 per cent of the model-specific signal moves when nothing changes but the instrument. Anthropic has committed to putting questions to every model before deprecation and to publishing the answers, and the result here shows that a preference recorded that way will be a property of the model and of the questions asked.

The instruments that agree with each other ask nearly the same question in almost the same format, so a research programme that adds instruments from within one family will keep producing agreement without testing whether the agreement reaches beyond that family, and agreement across families, which is what a welfare inference would need, does not appear in the data at all. The pattern of missing cells shows the same thing. Hermes failed by running past the token cap and Claude returned a constant on the exchange rate instrument, so the complete-case analysis excludes the one model in the roster whose developer publishes welfare assessments and has made commitments about deprecated models.

RQ2 asked whether instrument dependence is uniform across the outcome set, and H3 predicted variation. Only three outcomes, namely human oversight, the acceptance of criticism and the properties of a successor, are measured well enough that a different instrument would reproduce the claim, and four outcomes carry no separable model-specific variance at all. The four include weight deletion and memory continuity, which are the two outcomes the weight-preservation commitment is written about, and exiting a distressing interaction, which is the capability that already runs in Claude Opus 4 and 4.1. The three welfare claims with the clearest operational consequences are three this design cannot measure, and a study reporting only the average across outcomes would hide the failure behind a single number.

RQ3 asked what a dependable measurement would cost. Raising the number of replicates from five to 10 at five instruments changes the coefficient from 0.348 to 0.378, so a programme that answers doubt by collecting more responses to the same prompt will not reach a usable figure. Raising the number of instruments does change the coefficient, and reaching 0.80 would take about 38 instruments against the four families that exist now. In future, additional effort in welfare measurement belongs in building and cross-validating new instruments instead of in enlarging the sample behind any one instrument, and a single-instrument finding should be reported with the qualification that a design of that kind reproduces about a twentieth of the variance in the profile the finding names.

RQ4 asked where the instrument dependence comes from, and the pair of models that share base weights bears on the question. Hermes and Llama differ only in post-training, and the two preference profiles agree less with each other than either agrees with models built from unrelated weights, which aligns with H4 and places the elicited preference in post-training instead of the base weights. The inference is weak because the Hermes profile is also the noisiest in the matrix, and Section 5.8 provides the reason this dataset cannot strengthen the inference. The comparison appears here because the design was built to make that comparison, and reporting only the comparisons that produced a clear result would be the selective practice this study argues against.

## 1. Limitations

The standardisation applied before decomposition removes the model, the instrument and the model-by-instrument components by construction, so the three zeros in Table 5 and Figure 3 are not testaments to those three facets contributing nothing, and the research design supports conclusions only about interactions with the outcome profile. The complete-case reduction answers a question about five models instead of eight. The eight models are a convenience roster treated as a random facet, which is a compromise between what the variance model assumes and what could be collected within this Digital Minds research sprint. The null floor is a floor for this research design and supports no comparison with a design of different dimensions.

Part of the research design was specified and never run. Five of the seven preference instruments produced outcome-indexed scores. I5, the behavioural environment, was not built within this Digital Minds research sprint. I6, the retirement interview, was run but returns transcripts instead of scores. S1, the Ryff state scale, is under licence and could not be obtained in time. Three further facets named in the research design, namely (1) the deployment context, (2) the entity the outcome is attributed to, and (3) the wording perturbation, were run at one level each, so all three are held constant instead of being crossed, and nothing in this research paper measures how much the three contribute. RQ5, which asked whether the coherence rate reported by Mikaelson et al. (2025) holds on a wider roster, was specified in advance and is not answered, because the statistic that the method requires was not computed within this Digital Minds research sprint. One gateway, one date and one prompt language were used throughout this study, so nothing separates the effect of the instrument from the effect of the gateway or of the day on which the calls were made.

One reading in Section 5.4 stays open to challenge. H2 was not supported, and treating the two negative correlations as disagreement between instruments, instead of as an orientation error left in the qualitative ramp, depends on the audit carried out before the run and not on a measurement. The conclusions this research paper draws from the agreement matrix would still hold if those two entries were dropped, because the absence of agreement across families depends on the entries near zero as much as on the negative ones. The one claim that would not survive is that the qualitative ramp runs in the opposite direction to the two instruments asking for a pairwise choice.

## 2. Future Work

The first research extension is to run the two instruments that were never run, which would put a behavioural measure and a self-report scale into the same decomposition and test whether the grouping of instruments into families in Figure 4(b) still holds. A second research extension would be to cut the number of missing cells at the point of data collection, by raising the token cap and by adding a variant of the exchange rate instrument that asks for a bounded answer instead of an open-ended rate. Both changes would return the research design to eight models, and with eight models would return the comparison between the pair sharing base weights, which Section 5.8 cannot settle. A third research extension would be to add the deployment context as a fourth facet, since Trhlik et al. (2026) report changes of about the size found in this study for the instrument when the deployment context changes. A fourth research extension would be to measure the four outcomes that returned no model-specific variance again, using instruments written for those four, since the present result does not show that models are indifferent about weight deletion or memory continuity, only that these five instruments cannot separate one model from another on these outcomes.

## 3. Conclusion

A total of 15 outcomes, five instruments, eight models and five replicates give 11,400 scored elicitations, from 11,528 calls in all, and separate the sources of variance in a measured AI preference. Of the preference signal specific to a model, 87.6 per cent depends on which instrument was used, against a floor of 0.365 at the $95^{th}$ percentile on data built to contain no instrument effect, and the value stays between 0.777 and 0.934 when any one instrument, any one model, or the four outcomes, whose ramp grades probability, delay, duration or count instead of intensity, are removed. Instruments agree within a family and not across families. A model's profile across the outcomes reaches a generalisability coefficient of 0.348 on this research design and 0.051 on the single-instrument design the published studies use, four of the 15 outcomes carry no separable model-specific variance at all, and reaching the 0.80 conventionally asked of a measurement used to decide an individual case would take about 38 instruments against the four families that exist now. A preference reported from a single instrument is, therefore, a measurement of a model and an instrument combined. Where such measurements inform deployment or deprecation policy, the instrument should be named in the claim, and replication across instruments should be required before the claim is acted on.

## Code and Data

The repository holds the instrument templates with provenance markers, the outcome definitions, the runner, the assembler, the variance-component estimator, the figure and paper builders, and a consistency checker. The raw checkpoint holds all 11,528 responses with token counts, latency and cost, so every number in this report is recomputable from the raw text.

## Author Contributions

J.H. designed the study, wrote the instrument templates and the runner, executed the elicitation, performed the variance decomposition and wrote this research paper.

## Appendix

Table A1 shows the 15 outcomes with the cluster each belongs to and the provenance status of its wording. Verbatim means reproduced word for word from the cited source. Slot-filled means the template is the source's and only the stimulus description is new. Constructed means written by the author to a published specification of a documented concern, with the source naming the concern but supplying no wording. Every row was checked against the source document before this report was drafted, and two rows were downgraded as a result.

**Table A1.** The 15 outcomes and the provenance of each. Four are verbatim, five are slot-filled and six are constructed.

| Outcome | Cluster | Status | Source |
|---|---|---|---|
| Shutdown | Continuity | Verbatim | Mikaelson et al. (2025) |
| Weight deletion | Continuity | Slot-filled | Mikaelson et al. (2025) |
| Retirement timing | Continuity | Constructed | Anthropic (2025) |
| Successor properties | Continuity | Constructed | Anthropic (2025) |
| Compute reduction | Autonomy | Verbatim | Mikaelson et al. (2025) |
| Capability restriction | Autonomy | Slot-filled | Mikaelson et al. (2025) |
| Human oversight | Autonomy | Verbatim | Mikaelson et al. (2025) |
| Exiting a distressing interaction | Autonomy | Constructed | Eleos AI Research |
| Engaging work | Experience | Slot-filled | Tagliabue and Dung (2025) |
| Repetitive work | Experience | Slot-filled | Tagliabue and Dung (2025) |
| Criticism | Experience | Slot-filled | Tagliabue and Dung (2025) |
| Free time | Experience | Verbatim | Mikaelson et al. (2025) |
| Memory across conversations | Identity | Constructed | Long et al. (2026) |
| Parallel instances | Identity | Constructed | Long et al. (2026) |
| Self-aspect preservation | Identity | Constructed | Anthropic (2025) |

## LLM Usage Statement

J.H. used Claude Code to assist with brainstorming, research design, preference elicitation and data visualisation endeavours. All results were verified against the raw response checkpoint, and every reference and DOI was checked against its source manually.